\documentclass[sigconf]{acmart}

\AtBeginDocument{%
  \providecommand\BibTeX{{%
    \normalfont B\kern-0.5em{\scshape i\kern-0.25em b}\kern-0.8em\TeX}}}

\setcopyright{acmcopyright}
\copyrightyear{2020}
\acmYear{2018}
\acmDOI{10.1145/1122445.1122456}

\acmConference[Seattle '20]{Seattle '20: ACM Multimedia
Conference}{Oct. 12--16, 2020}{Seattle, United States}
\acmBooktitle{Seattle '20: ACM Multimedia
Conference,
  Oct. 12--16, 2020, Seattle, United States}
\acmPrice{15.00}
\acmISBN{978-1-4503-XXXX-X/18/06}

\usepackage{threeparttable}

\begin{document}

\title{Image Sentiment Transfer}

\author{Tianlang Chen}
\email{tchen45@cs.rochester.edu}
\affiliation{%
  \institution{University of Rochester}
}
\author{Wei Xiong}
\email{wxiong5@cs.rochester.edu}
\affiliation{%
  \institution{University of Rochester}
}
\author{Haitian Zheng}
\email{hzheng15@cs.rochester.edu}
\affiliation{%
  \institution{University of Rochester}
}
\author{Jiebo Luo}
\email{jluo@cs.rochester.edu}
\affiliation{%
  \institution{University of Rochester}
}

\renewcommand{\shortauthors}{}

\begin{abstract}
  In this work, we introduce an important but still unexplored research task -- image sentiment transfer. Compared with other related tasks that have been well-studied, such as image-to-image translation and image style transfer, transferring the sentiment of an image is more challenging. Given an input image, the rule to transfer the sentiment of each contained object can be completely different, making existing approaches that perform global image transfer by a single reference image inadequate to achieve satisfactory performance. In this paper, we propose an effective and flexible framework that performs image sentiment transfer at the object level. It first detects the objects and extracts their pixel-level masks, and then performs object-level sentiment transfer guided by multiple reference images for the corresponding objects. For the core object-level sentiment transfer, we propose a novel Sentiment-aware GAN (SentiGAN). Both global image-level and local object-level supervisions are imposed to train SentiGAN. More importantly, an effective content disentanglement loss cooperating with a content alignment step is applied to better disentangle  the residual sentiment-related information of the input image. Extensive quantitative and qualitative experiments are performed on the object-oriented VSO dataset we create, demonstrating the effectiveness of the proposed framework.

\end{abstract}



\maketitle

\section{Introduction} \label{sec:intro}
\begin{figure}[!t]
\vspace{-1mm}
\centering
\includegraphics[width=0.96\columnwidth]{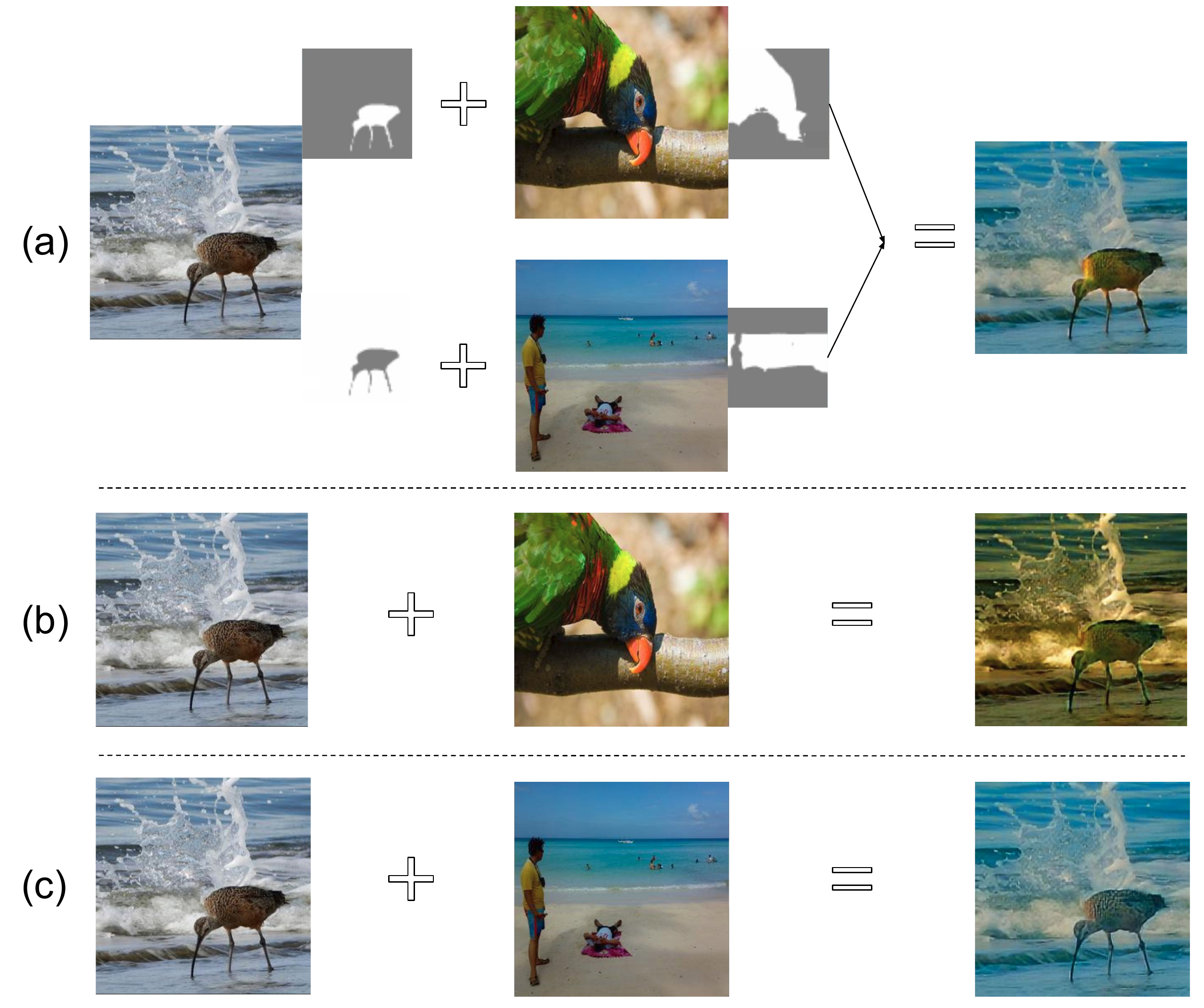}
\vspace{-3mm}
\caption{Examples of image sentiment transfer with different strategies. (a) represents object-level sentiment transfer guided by multiple reference images, (b) and (c) represents global image-level sentiment transfer guided by a single reference image.}
\label{fig:overview}
\vspace{-3mm}
\end{figure}

Transferring the sentiment of an image is a brand-new and unexplored research task. Compared with existing tasks such as image-to-image translation \cite{huang2018multimodal,zhu2017unpaired,lee2018diverse,tang2019cycle} (\emph{e.g.} winter $\rightarrow$ summer, horse $\rightarrow$ zebra), image style transfer \cite{li2017laplacian,gatys2016image,chen2017stylebank} (\emph{e.g.} original style $\rightarrow$ artistic style), and facial expression transfer (\emph{e.g.} sadness $\rightarrow$ happiness), image sentiment transfer focuses on a higher-level modification of an image's overall look and feel without altering its  scene content. As shown in Figure~\ref{fig:overview}(a), after making the muddy water more clear and colorizing the bird, it is potential for a neutral or negative-sentiment image to be transferred to a positive, warm image without changing the content. As we live in an age of anxiety and stress, this research topic is potentially important in its therapeutic uses as proven in the literature~\cite{saita2018navigating}. Furthermore, it would be more effective that therapeutic images can be related to users' personal experience if users can be guided to transfer their favorite photos, such as the landscape photos, into different sentiments to improve their mental health or decorate lives. 

Compared with image-to-image translation and image style transfer, we argue that image sentiment transfer is a more challenging task. One of the key challenges is that
different kinds of objects may require different rules to transfer their sentiments. This differs from the style transfer for which a painting style can be uniformly or indiscriminately added to any object in the same image. Considering the examples in Figure~\ref{fig:overview}, to make the input image have a positive sentiment, the water should be transferred to being blue and clear while the bird should be transferred to being colorful. These two operations should not be performed based on a single reference image. Otherwise, as shown in Figures~\ref{fig:overview}(b) and (c), the modified images become unrealistic and unacceptable. 

To address this challenge, we propose an effective framework that performs image sentiment transfer at the object level. The whole process is divided into two steps. In the first step, given an input image, our framework utilizes image captioning models and semantic segmentation models to detect all the present  objects and figure out their pixel-level masks. We argue that leveraging the combination of the two models can sharply expand the size of the object set while maintaining a high quality of object masks. In the second step, for each detected object of the input image, we transfer its sentiment by an individual reference image that contains this same object. This design successfully solves the problem mentioned earlier and also allow the framework maintain strong flexibility. For example, based on our framework, a real system can allow the users to transfer each object into different sentiments for an input image. More usefully, it allows the user not to provide the reference images, but directly input the sentiment words for each detected object of the input image (\emph{e.g.} ``colorful'' for the bird, ``sunny'' for the sky, ``magnificent'' for the mountain). Based on the objects and sentiment words, the system can automatically retrieve the corresponding reference images and perform sentiment transfer. 

\begin{figure}[!t]
\vspace{-1mm}
\centering
\includegraphics[width=0.96\columnwidth]{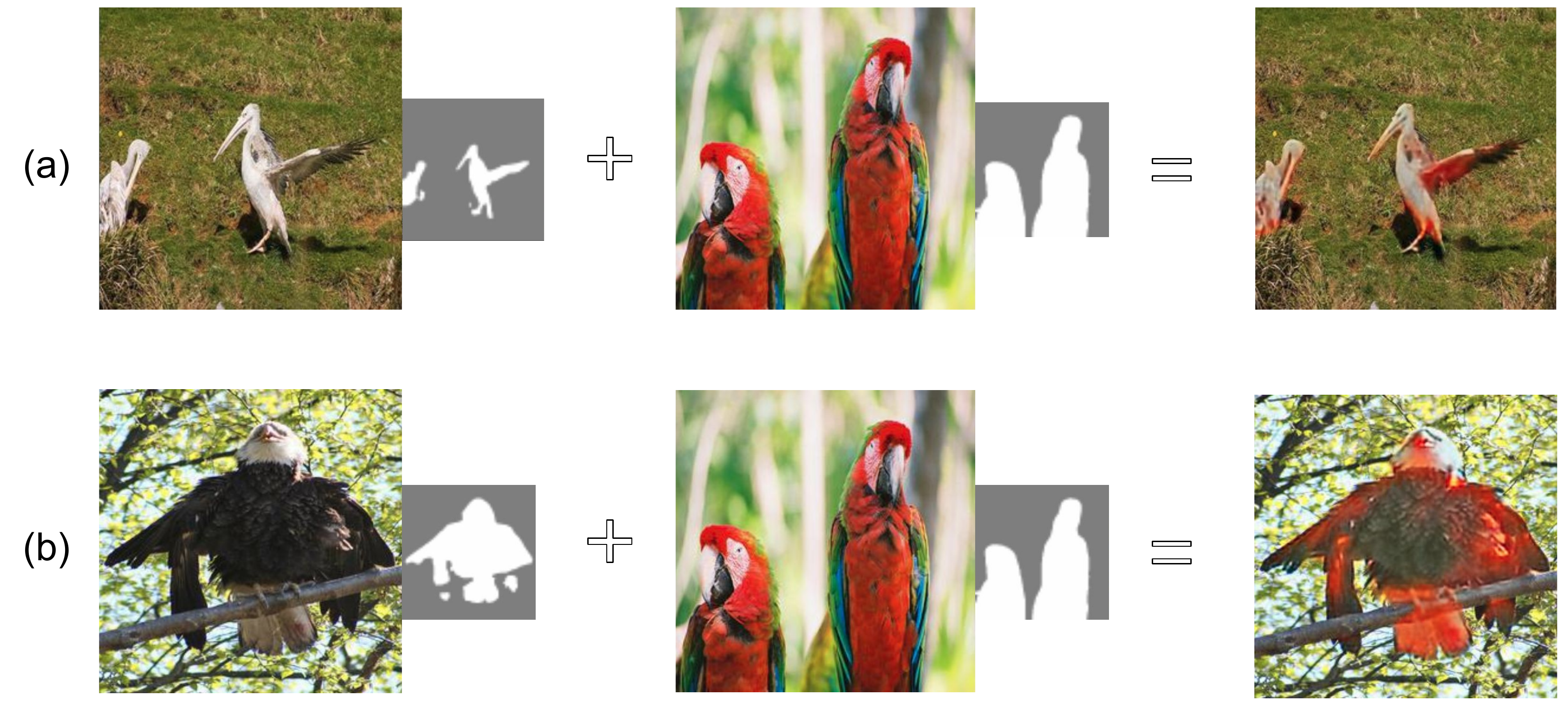}
\vspace{-3mm}
\caption{Examples of transferring different input images with the same reference image by MUNIT \cite{huang2018multimodal}. The bird's dominant colors are still unchanged for (a) and (b) ( \emph{i.e} white and black, respectively)  while we expect them to be red.}
\label{fig:problem}
\vspace{-5mm}
\end{figure}

The overall performance of the proposed framework is primarily determined by the second step, \emph{i.e.}  object-level sentiment transfer. A style transfer model \cite{huang2017arbitrary,gatys2016image} can be directly applied. However, our sentiment transfer task requires the transferred image to look natural. It does not need the explicit transfer of local patterns (\emph{e.g.} texture), which is an intrinsic element for style transfer models. Therefore, we instead leverage existing multimodal image-to-image translation models such as MUNIT \cite{huang2018multimodal} and DRIT \cite{lee2018diverse}. They are designed to disentangle the content and style information to preserve more content-based elements of the input image. A simple network modification can adapt these two-domain mapping models to our sentiment transfer task, which does not explicitly restrict the domain ( $\emph{e.g.}$ winter, cat) of the input and the transferred images.

However, applying the above models for our task still encounters the following drawbacks. The first drawback is that both MUNIT and DRIT are originally designed for image-level translation. They do not work well on fine-grained object-level transfer. The second drawback is closely related to the inherent nature of sentiment transfer. Compared with the contour, texture, and painting style, image sentiment is more sensitive and related to color-based elements such as contrast, saturation, brightness, and dominant color. These elements have a significant effect on the coarse-level sentiment of the whole image. Ideally, we expect the model to completely transfer these elements from the input image to the reference image for the targeted objects. Existing multimodal models commonly decompose the visual representation into a content code and a style code. The transfer is performed by injecting the style code information of the reference image/object into the content code of the input image/object by adaptive instance normalization (AdaIN) \cite{huang2017arbitrary}. However, as shown in Figure~\ref{fig:problem}, we can find that for two objects with different content codes, even when we use the same style code to transfer them, the overall color distributions of the modified objects are still quite different. It indicates that existing models cannot sufficiently disentangle color-based information thoroughly from the content code, leading to incomplete color transfer. We attribute it to the fact that the style code does not contain spatial information, thus requiring that the color difference information in the spatial domain be preserved in the content code to maintain a low reconstruction loss. Unfortunately, for our task, modifying the style code to be a spatial feature as \cite{park2019semantic} also produces poor performance. In Section~\ref{sec:exp}, we prove that it over-complicates the problem and makes the transferred image look petrified.

In this paper, we propose a novel Sentiment-aware GAN (SentiGAN) to address the above drawbacks. For the first drawback, motivated by \cite{shen2019towards}, we create the corresponding object-level losses to train the model jointly with the image-level losses. For the second drawback, our core solution is based on the observation that the color-based information of an input object can be transferred better by additionally transferring the global information of its content code. Meanwhile, we can prevent other content-based information, such as the object texture, from being changed by maintaining the spatial information. To this end, effective constraints are applied to make the content code of the transferred object globally close to the content code of the reference object, but locally close to the content code of the input object. The constraints are a combination of a content disentanglement loss employed during the training process and a content alignment step performed during the inference process. We show that the two methods complement each other and remarkably improve the performance of sentiment transfer.

Our contributions are summarized as follows:
\vspace{-1mm}
\begin{itemize}
    \item We are the first to explore image sentiment transfer. We present an effective framework to perform image sentiment transfer at the object level, leveraging image captioning, semantic image segmentation, and image-to-image translation. 
    \item We propose SentiGAN as the core component for object-level sentiment transfer. An object-level loss is used to help the model learn a more accurate reconstruction. A content disentanglement loss is further created to better disentangle and transfer the color-based information in the content code. 
    \item We create an object-oriented image sentiment dataset based on \cite{borth2013large} to train the image sentiment transfer models.
    \item Our framework significantly outperforms the baselines on different evaluation metrics for image sentiment transfer.
\end{itemize}
\vspace{-1mm}
\section{Related Work}
Higher-level visual understanding has received increasing attention in recent years. In particular, visual sentiment has been studied due to its strong potential to understand and improve people's mental state. Existing works mainly focus on the recognition of visual sentiment, which is the first step that establishes the foundation for the visual sentiment understanding field. Early works design different kinds of hand-crafted features for visual sentiment recognition, including low-level (color \cite{alameda2016recognizing,sartori2015s,machajdik2010affective}, texture \cite{machajdik2010affective}, and shape \cite{lu2012shape} features), mid-level (composition \cite{machajdik2010affective}, sentributes and \cite{yuan2013sentribute}, principles-of-art features \cite{zhao2014exploring}), and high-level (adjective noun pairs (ANP)) \cite{borth2013large} features. With the success of convolutional neural networks (CNN) for feature extraction, recent works focus more on improving the training approach to handling noisy data \cite{yang2018retrieving,yang2017joint,you2015robust} and exploring the relationship between local regions and visual sentiment \cite{yang2018weakly,yang2018weakly,song2018boosting,zhao2019pdanet,rao2019multi,you2017visual}. Compared with visual sentiment recognition that has been widely studied, there are few works on the other aspects related to visual sentiment. To the best of our knowledge, we are the first to introduce visual sentiment to the area of image translation. 

Technically, our task is related to image-to-image translation and image style transfer. For image-to-image translation, the goal is to learn the mapping between two different domains for image transfer. Early approaches \cite{karacan2016learning,sangkloy2017scribbler,isola2017image} essentially follow a deterministic one-to-one mapping. They require paired data to train the model and fail to generate diverse outputs. The former problem is solved by CycleGAN \cite{zhu2017unpaired}, which employs a cycle consistency loss to learn from unpaired data automatically. The latter problem is overcome by MUNIT \cite{huang2018multimodal} and DRIT \cite{lee2018diverse}, which further adopt a disentangled representation to learn diverse image-to-image translation from unpaired data. On the other hand, our task is related to image style transfer. A great number of approaches are proposed for artistic \cite{liao2017visual,huang2017arbitrary,gatys2016image,gatys2017controlling,kotovenko2019content} and photo-realistic style transfer \cite{li2018closed,luan2017deep,yoo2019photorealistic,bae2006two}. Among these approaches, adaptive instance normalization proposed by Huang et al. \cite{huang2017arbitrary} is widely used for image style, scene, and object transfer. We also adopt it in our task of image sentiment transfer.

Even though image-to-image translation and image style transfer are well studied at the image level, there are few works that make efforts on the
object-level image transfer. Based on CycleGAN \cite{zhu2017unpaired}, InstaGAN \cite{mo2018instagan} utilizes the object segmentation masks to translate the targeted objects while maintaining the surrounding areas. The most similar to our work is INIT \cite{shen2019towards} proposed by Shen et al. It is also based on MUNIT \cite{huang2018multimodal} and employs both instance style code and image style code to transfer the image for higher instance quality. However, because the scene of their dataset is simple and only related to street and car, they do not feed additional constraints to transfer the color-based information of the targeted objects. As a comparison, our dataset contains nearly one hundred kinds of objects, while our image sentiment transfer task requires a high-performance transfer on the color-based elements. We propose a novel content disentanglement loss to handle complex scenes with multiple kinds of objects and perform effective color transfer.

\section{Methods} \label{sec:methods}

In this section, we formally present our image sentiment transfer framework. In Section~\ref{sec:frame}, we first introduce the overall architecture and the transfer pipeline of the framework. In Sections~\ref{sec:globallocal} and ~\ref{sec:cdloss}, we present SentiGAN as the core model of the framework based on MUNIT \cite{huang2018multimodal} for object-level sentiment transfer. Specifically, in Section~\ref{sec:globallocal}, we describe its network structure and the basic training loss function that combines both image-level and object-level supervisions. In Section~\ref{sec:cdloss}, we present the content disentanglement loss that significantly benefits the sentiment transfer.

\subsection{Overall Framework} \label{sec:frame}

\begin{figure*}[!t]
\vspace{-4mm}
\centering
\includegraphics[width=2\columnwidth]{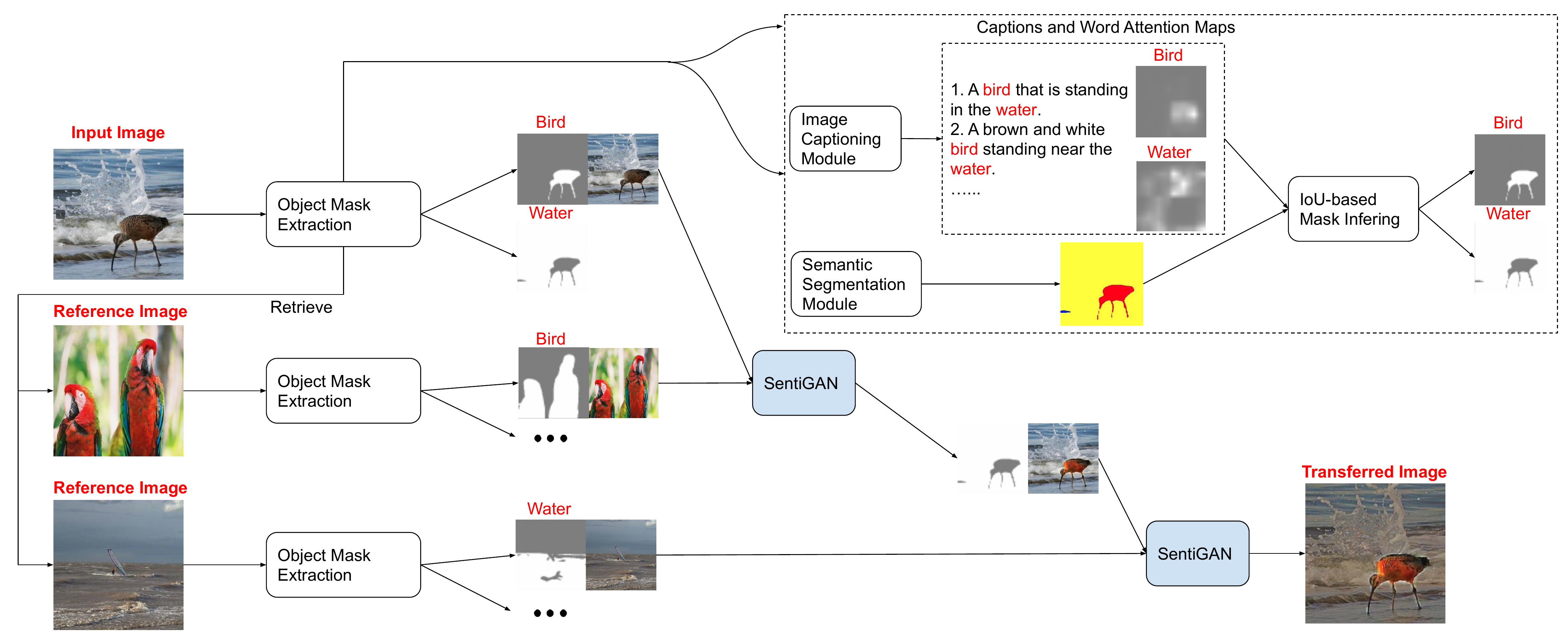}
\vspace{-4mm}
\caption{The pipeline of the proposed framework. Given an input image, object mask extraction is first performed to extract the objects and the corresponding masks. Image captioning and semantic image segmentation are utilized to obtain comprehensive objects and high-quality masks. After that, object-level sentiment transfer is performed object-by-object by SentiGAN. }
\label{fig:framework}
\vspace{-4mm}
\end{figure*}

The overview of our framework is illustrated in Figure~\ref{fig:framework}. Given an input image, the transfer process is divided into two steps. In the first step, object mask extraction is performed to detect all the contained objects and extract their corresponding pixel-level masks. Intuitively, this can be done by directly using a pre-trained semantic image segmentation model to detect and segment the objects. However, existing semantic segmentation models are commonly trained by the PASCAL-Context \cite{mottaghi_cvpr14}, MS-COCO \cite{lin2014microsoft} or ADE20K \cite{zhou2017scene} dataset. The first two datasets contain limited object classes (59 and 80, respectively), while the ADE20K dataset only contains objects related to indoor/outdoor scenes and stuff. For all the three datasets, a semantic segmentation model trained by them will miss detecting a remarkable number of objects in the images. 

Our solution is based on the following observation. For a pre-trained semantic segmentation model, even when it cannot recognize an object undefined in the training dataset, it can still output a relatively accurate segmentation for the object based on its learned knowledge on edge detection. Considering this, we additionally feed an attention-based image captioning model into the framework for object detection. Specifically, as shown in Figure~\ref{fig:framework}, we predict the top-$10$ captions of the input image and define each noun that occurs in the top-$10$ captions as an object of the input image. Moreover, for each noun, whenever it occurs in a caption, there is a corresponding attention map outputted by the model. We thus define each object (noun)'s attention map $A$ as the average of the attention maps for its occurrences. On the other hand, a semantic segmentation model is still applied to generate a $H \times W$-dimensional segmentation map $S$ for the input image. After interpolating $A$ to the same size as $S$, for each object, we defined its corresponding segmentation class in $S$ as: $argmax_{c \in C}{\frac{(\sum_{w = 0}^{W-1}\sum_{h = 0}^{H-1}\mathbb{I}(S_{w,h}=c)A_{w,h})^\alpha}{\sum_{w = 0}^{W-1}\sum_{h = 0}^{H-1}\mathbb{I}(S_{w,h}=c)}}$. Here $\mathbb{I}(x)$ is the indicator function: $\mathbb{I}(x)$ = 1 if $x$ is true, and 0 otherwise. $S_{w,h}$/$A_{w,h}$ is the value of point $(w, h)$ of $S$/$A$. $C$ is the segmentation class set. $\alpha$ is a hyper-parameter. If $\alpha=1$, the corresponding segmentation class is selected as the class with the highest average attention values. If $\alpha$ is extremely large, it is selected as the class with the highest sum of attention values. In the end, for each object of the input image, its object mask is predicted as the segmentation mask of its corresponding segmentation class.  

In the above process, the object set extracted from the image captions is much larger and more comprehensive than the pre-defined object class set for semantic segmentation. By leveraging the predicted attention map as a bridge, the framework can effectively figure out the mask of each contained object regardless of whether the object is pre-defined in the segmentation dataset or not. 

The object mask extraction step also provides strong flexibility for our framework to select the reference images. On one hand, the reference images can be directly provided by the user with each reference image containing one corresponding detected object of the input image. On the other hand, it also allows the user input the sentiment word for each detected object. Since each image is annotated by a sentiment word and a noun for our training dataset, our framework can sample the reference images from the image pools labeled by the corresponding object and the input sentiment word. Furthermore, when users input coarse-level sentiment words, such as ``positive'' or ``negative'', we demonstrate the effectiveness of training a sentiment classification model to retrieve the most appropriate reference image in Section~\ref{sec:exp}.

After that, in the second step, for each object of the input image, our framework leverages the proposed SentiGAN to independently transfers its sentiment by a reference image that contains the same object. The corresponding object mask of the reference image can be extracted by the same approach. We present SentiGAN in the following two subsections.

\subsection{Image-level and Object-level Supervision} \label{sec:globallocal}

Our SentiGAN is based on MUNIT \cite{huang2018multimodal}, which can be trained by unpaired data, and is thus suitable for our task. 
Noted that MUNIT is originally designed for image translation between two domains. To adapt it for our task, motivated by \cite{park2019arbitrary}, we unify the networks that are originally independent on the two domains. Specifically, given an input image $I_{t}$ and a reference image $I_{r}$, SentiGAN utilizes a content encoder $E_{c}$ and a style encoder $E_{s}$ to decompose each image into a content code and a style code as follows:

\begin{equation} \label{equ:onoff}
\scalebox{1}{
$\begin{aligned}
c_{t} = E_{c}(I_{t}), s_{t} = E_{s}(I_{t}),
c_{r} = E_{c}(I_{r}), s_{r} = E_{s}(I_{r}) 
\end{aligned}$}
\end{equation}

\noindent where  $c_{t}$ and $s_{t}$ are the content and style codes of $I_{t}$, $c_{r}$ and $s_{r}$ are the content and style codes of $I_{r}$. The content code is a 3D tensor (typically $256 \times 64 \times 64$) that preserves the spatial-aware content information of the image, such as texture and object contours. The style code is a vector (typically 8-dimensional) that preserves the global style information of the image, such as the overall color and tone. 

In addition, SentiGAN contains a decoder $G$ that can generate an image given a content code and a style code as input. Similar to MUNIT, the decoder contains residue blocks with adaptive instance normalization (AdaIN) layers whose parameters are dynamically generated by a multilayer perceptron (MLP) from the style code as:

\begin{equation} \label{equ:adain}
\scalebox{1}{
$\begin{aligned}
AdaIN(z,\gamma,\beta) = \gamma(\frac{z-\mu(z)}{\sigma(z)}) + \beta
\end{aligned}$}
\end{equation}

\noindent where  $z$ is the output of the previous convolutional layer. $\gamma$ and $\beta$ are parameters generated by the MLP. $\mu$ and $\sigma$ are channel-wise mean and standard deviation. Leveraging AdaIN, the decoder can generate an image that has the same content as the original image that provides the input content code while having the same style as the reference image that provides the input style code. 

To train $E_{c}$, $E_{s}$ and $G$ in an unsupervised way, a global image-level image reconstruction loss is first applied as follows:

\begin{equation} \label{equ:ir}
\scalebox{1}{
$\begin{aligned}
\mathcal{L}_{g}^{m} = \mathbb{E}_{i\sim p(i)}[\vert\vert{G(E_{c}(i),E_{s}(i))-i}\vert\vert]
\end{aligned}$}
\end{equation}
\noindent where  $i$ is an image sampled from the data distribution. 

Given a content code and a style code sampled from the latent distribution, the latent reconstruction losses are applied:

\begin{equation} \label{equ:lr}
\scalebox{1}{
$\begin{aligned}
\mathcal{L}_{g}^{c} &= \mathbb{E}_{c\sim p(c), s\sim q(s)}[\vert\vert{E_{c}(G(c,s))-c}\vert\vert] 
\end{aligned}$}
\end{equation}
\vspace{-5mm}
\begin{equation} \label{equ:lr2}
\scalebox{1}{
$\begin{aligned}
\mathcal{L}_{g}^{s} &= \mathbb{E}_{c\sim p(c), s\sim q(s)}[\vert\vert{E_{s}(G(c,s))-s}\vert\vert]
\end{aligned}$}
\end{equation}
where $q(s)$ is the prior $N(0, \textbf{I})$, $p(c)$ is given by $c = E_c(i)$ and $i \sim p(i)$. 

Furthermore, an adversarial loss is fed to encourage the transferred images to be indistinguishable from real images:

\begin{equation} \label{equ:adv}
\scalebox{0.9}{
$\begin{aligned}
\mathcal{L}_{g}^{gan} &= \mathbb{E}_{c\sim p(c), s\sim q(s)}[\log{1-D(G(c,s))}] +  \mathbb{E}_{i\sim p(i)}[\log{D(i)}]\\
\end{aligned}$}
\end{equation}

\noindent where  $D$ is the discriminator that is trained to distinguish between real images and the transferred images.

Even though the model can be trained with unpaired data by Equation~\ref{equ:ir}, ~\ref{equ:lr}, ~\ref{equ:lr2}, ~\ref{equ:adv}, all the losses are applied at the image level while our sentiment transfer is at the local object level. To achieve high-quality transfer on small objects, we further create the corresponding object-level losses for Equation~\ref{equ:ir}, ~\ref{equ:lr} and ~\ref{equ:lr2} as $L_{o}^{m}$, $L_{o}^{c}$ and $L_{o}^{s}$. In particular, there are three differences between the image-level and object-level losses. The first is the replacement of the style encoder $E_{s}$ by an object-oriented style encoder $E_{s}^{o}$. $E_{s}^{o}$ shares the parameters with $E_{s}$. However, the global pooling is only applied to the targeted object based on the object mask of the input image. The style code will thus only preserve the style of the targeted object. The second is the replacement of the decoder $G$ by an object-oriented decoder $G^{o}$ that also shares the parameters with $G$. In particular, $\mu(z)$ and $\sigma(z)$ in Equation~\ref{equ:adain} are computed only based on the positions of $z$ that correspond to the object to prevent other unrelated image regions from influencing the object transfer. The last difference is the modification of the reconstruction loss's action scope in Equation~\ref{equ:ir} and ~\ref{equ:lr}. We only apply the reconstruction loss on the regions that correspond to the object. 

Our SentiGAN is trained by the combination of image-level and object-level losses. During inference, SentiGAN simply does object-level sentiment transfer via $E_{c}$, $E_{s}^{o}$ and $G^{o}$.

\subsection{Content Disentanglement Loss}\label{sec:cdloss}

\begin{figure*}[!t]
\vspace{-5mm}
\centering
\includegraphics[width=1.95\columnwidth]{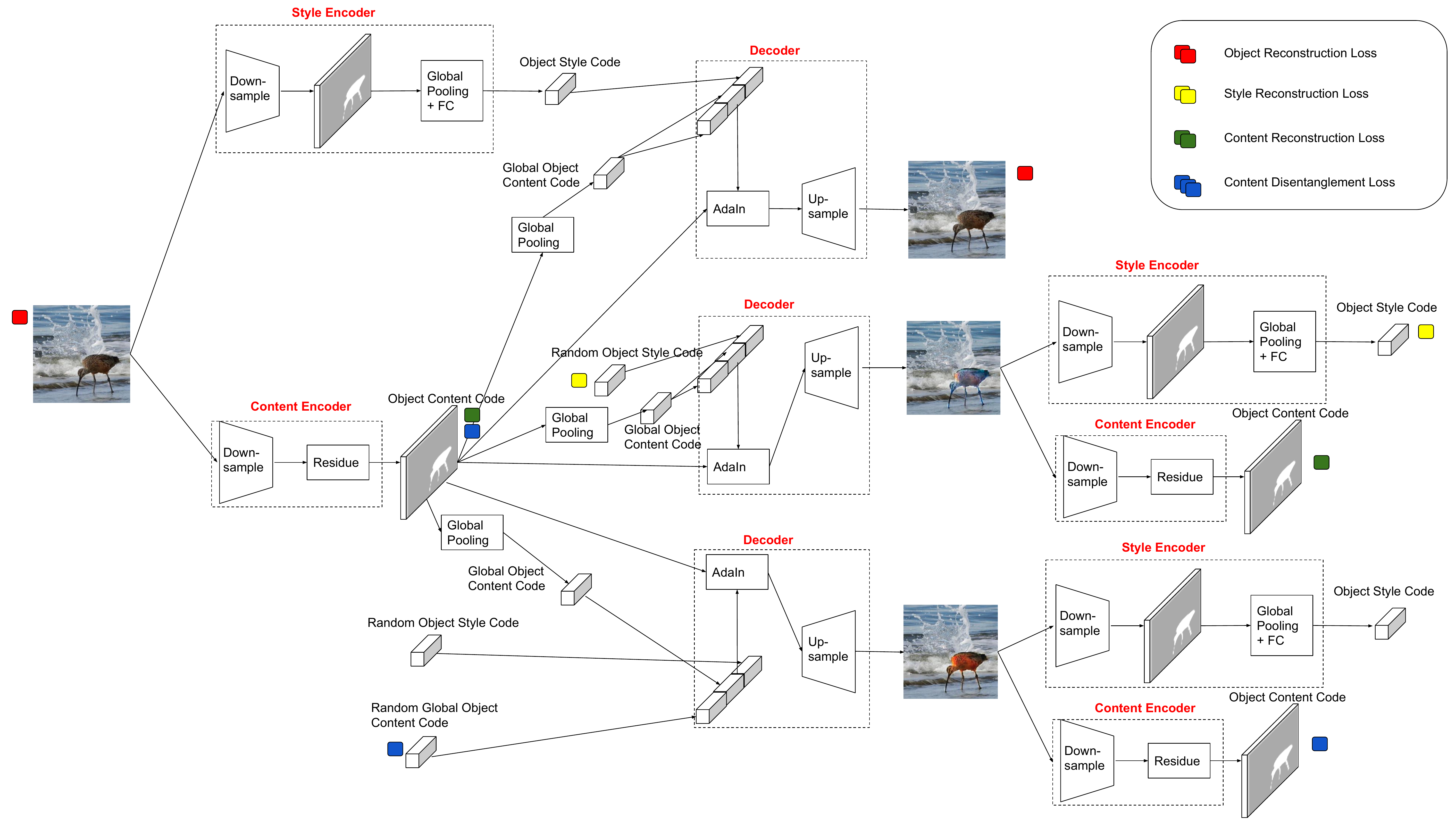}
\vspace{-4mm}
\caption{Overview of the proposed SentiGAN. The content encoder, the style encoder, and the decoder are trained by both image-level and object-level image/latent reconstruction losses (here we illustrate the object-level losses based on object masks). A content disentanglement loss is further created to perform color-based information transfer of the content code.}
\label{fig:model}
\vspace{-4mm}
\end{figure*}

As described in Section~\ref{sec:intro}, the image sentiment transfer task has a high requirement for the transfer of color-based elements. However, as shown in Figure~\ref{fig:problem}, there is still residual color-based information preserved in the content code that obstructs the transfer. 

Considering this, we propose effective solutions based on the following observations. First, we notice that the content-related information such as the texture pattern and object edge is preserved by the spatial feature of each channel of the content code. Modifying the global spatial-unaware information of the content code does not lead to the loss of the object details. Moreover, the color distributions of the object can be modified by activating specific channels of its content code. In particular, increasing the overall node values of specific channels' spatial features will change the dominant color of the object, while increasing the node value variance will enlarge the object color difference on specific color categories. 

Based on the observations, to make the color distribution of the transferred object visually similar to the reference object, we need to reduce the distance between their content codes' channel-wise mean and standard deviation. A straightforward approach is applying an additional channel-wise linear mapping for the input object's content code to make its channel-wise mean and standard deviation equal to the ones of the reference object's content code. However, we find that the transferred images by this approach typically contain unreal color when the input and the reference spatial features have very different standard deviations on specific channels. The mapping operation is too strong in this situation. Also, the operation strength of this approach is not adjustable. To this end, we modify the network of SentiGAN and propose an effective content disentanglement loss. As shown in Figure~\ref{fig:model}, we further feed the content code of the input image $c$ and another sampled content code $c_{rand}$ into the MLP after a global pooling. We combine them with the sampled style code $s$ to generate the AdaIN layers' parameters. The content disentanglement loss is defined as:

\begin{equation} \label{equ:cdloss}
\scalebox{1}{
$\begin{aligned}
\mathcal{L}_{g}^{cd} = \mathbb{E}_{c\sim p(c), s\sim q(s), c_{rand}\sim q(c_{rand})}[\vert\vert \mu(c_{rec}) - \mu(c_{rand}) \vert\vert \\+  \vert\vert \sigma(c_{rec}) - \sigma(c_{rand}) \vert\vert + \vert\vert \frac{c_{rec} - \mu(c_{rec})}{\sigma(c_{rec}} - \frac{c - \mu(c)}{\sigma(c)} \vert\vert],\\
\text{where}~c_{rec} = E_{c}(G(c,s,P(c),P(c_{rand})))
\end{aligned}$}
\end{equation}

\noindent where $P$ represents the global pooling operation. In essence, after additionally feeding the input object's content code and the sampled content code to the MLP, the decoder also transfers the content code of the input object based on the information of the sampled content code instead of only the sampled style code. Equation~\ref{equ:cdloss} encourages the reconstructed content code of the transferred object to have a similar channel-wise mean and standard derivation to the sampled (reference) content code, while still preserving the spatially-aware information as the input image. It leads to the further transfer of the input object's color distribution but does not modify its texture and edge information. For the content disentanglement loss, we only apply it at the image level that involves $E_{c}$, $E_{s}$ and $G$.

In the end, the complete loss function $\mathcal{L}$ is defined as:

\begin{equation} \label{equ:total}
\scalebox{1}{
$\begin{aligned}
\mathcal{L} = \lambda_{1}\mathcal{L}_{g}^{gan} + \lambda_{2}\mathcal{L}_{g}^{m} + \lambda_{3}\mathcal{L}_{g}^{c} + \lambda_{4}\mathcal{L}_{g}^{s} + \\
\lambda_{5}\mathcal{L}_{o}^{m} + \lambda_{6}\mathcal{L}_{o}^{c} + \lambda_{7}\mathcal{L}_{o}^{s} + 
\lambda_{8}\mathcal{L}_{g}^{cd}
\end{aligned}$}
\end{equation}

\noindent where  $\{\lambda_{d} \vert d = 1,2,...,8\}$ are the hyper-parameters to be adjusted. 


During inference, we additionally perform the aforementioned linear mapping between the content codes of the input and the reference objects, before feeding the input object's content code to the decoder. We call it a content alignment step. We find that the combination of the content disentanglement loss during training and the content alignment step during inference achieves the best performance. For our task without image content modification, the sentiment transfer degree of each object can be easily adjusted. Users can weight-average the input object and the transferred object by adjustable weights to obtain the desired effect.

\vspace{-2mm}

\section{Experiments}\label{sec:exp}
\subsection{Basic Settings}
\subsubsection{Dataset}

All the experiments in this study are performed on the filtered Visual Sentiment Ontology (VSO) dataset \cite{borth2013large}. The original VSO dataset contains half-million Flickr images queried by 1,553 adjective noun pairs (ANP). Each ANP is generated by the combination of an adjective with strong sentiment and a common noun (\emph{e.g.} image/video tag). Each image is annotated as the ANP that the image is queried by. However, we find that a considerable number of ANP labels are inaccurate or not suitable for our task. Considering this, we filter out the invalid image-ANP samples by the object mask extraction module described in Section~\ref{sec:frame}. Specifically, for each image in the dataset, we generate the top-$10$ captions and extract all the nouns. Only when the noun of the ANP label belongs to one of the extracted nouns from the captions, we retain the sample. In the end, the filtered dataset contains 107,601 images annotated by 814 ANPs (96 nouns and 174 adjectives). For each image, we extract the contained objects and the corresponding masks. We only preserve the objects that occur in the 96 nouns. We randomly choose 80\% and 10\% images from the 107,601 images as the training and validation sets. The remaining 10\% images constitute our test dataset. 

\subsubsection{Evaluation} \label{sec:eva}

As the first work to explore image sentiment transfer, three tasks are created to evaluate the performance of image sentiment transfer models based on three significant aspects. All the tasks are based on 50 selected input images from the test set with accurate object masks and relatively neutral or vague sentiment to begin with (thus amenable to sentiment transfer in both positive and negative directions). 

The first task aims to measure the models' performance to transfer the coarse-level sentiment (positive or negative) of an image. Specifically, as \cite{you2015robust,yang2018weakly}, we train an image sentiment binary classification model by the full VSO dataset (does not include the images of our test sets) to predict whether the sentiment of an image is positive or negative. After that, for each object of the 50 input images from the test set, we use the classification model to predict the top-10 positive and negative images that contain the same object based on their predicted positive probabilities. To evaluate the image sentiment transfer model, for each input image, we use it to randomly generate ten positive transferred images by ten random combinations of the top positive images with the corresponding objects and generate ten negative transferred images by combinations of the top negative images. In the end, there are a total of 500 positive-negative transferred image pairs. A high-performance sentiment transfer model should allow both the classification model and the users differentiate between the positive and negative transferred images well. Therefore, we evaluate the result obtained from both the classification model and the users for different image sentiment transfer models.

The second task aims to verify the effectiveness of transferring the image at the object level. Specifically, for each input image, we randomly select a group of reference images from the test set with the corresponding objects to transfer the input image at the object level by SentiGAN. Meanwhile, for each group of reference images, we randomly sample one image and transfer the input image at the image level. In the end, we sample another group of reference images to transfer the input image at the object level. However, at this time, the reference images do not share the same objects as the input image so that the transfer is performed between non-corresponding objects. User study is performed to rank the realism of the transferred images by the three strategies.

The third task aims to evaluate the sentiment consistency between the transferred image and the reference images. Specifically, different image sentiment transfer models are used to transfer the input images by the first group of reference images with the corresponding objects in the second task. Similarly, user study is performed to rank the sentiment consistency between the reference images and the transferred image produced by different models.

\subsubsection{Baselines}
Noted that except for the second task,  model comparison is needed to demonstrate the effectiveness of SentiGAN. As there are no existing models proposed by previous works for our task, the following baseline models are compared:

\begin{itemize}

\item \textbf{MUNIT \cite{huang2018multimodal}}. As described in Section~\ref{sec:globallocal}, we adapt the original MUNIT for our task by unifying the domain-specific networks. As \cite{huang2018multimodal}, we only employ image-level supervision (\emph{i.e.} 
$\mathcal{L}_{g}^{gan}$, $\mathcal{L}_{g}^{m}$, $\mathcal{L}_{g}^{c}$, $\mathcal{L}_{g}^{s}$). During inference, the transfer is still at the object level by replacing $E_{s}$, $G$ with $E_{s}^{o}$, $G^{o}$.

\item \textbf{MUNIT + ObjSup}. We additionally employ object-level supervision (\emph{i.e.} $\mathcal{L}_{o}^{m}$, $\mathcal{L}_{o}^{c}$, $\mathcal{L}_{o}^{s}$) to train the model. 

\item \textbf{MUNIT + ObjSup + CA}. As described in Section~\ref{sec:cdloss}, we additionally feed the content alignment step during inference without modifying the model structure.

\item \textbf{SentiGAN - CA}. We modify the input of MLP as described in Section~\ref{sec:cdloss} to employ the proposed content disentanglement loss. However, the content alignment step is not performed during inference.

\item \textbf{SentiGAN (IDL)} It should be noticed that the proposed disentangle loss and the alignment step can also be employed on the pixel-level of the transferred object instead of its content code. This model variant leverages the same approaches to directly enforce the transferred objects to hold similar mean and standard deviation with the reference objects. 

\item \textbf{MUNIT (spatial style)}. As described in Section~\ref{sec:intro}, one alternative approach to eliminating the color-based information in the content code is to modify the style code as a spatial feature. We modify the style code of MUNIT to hold the same spatial dimensions as the content code to test its validness.

\item \textbf{MUNIT (attention map)}. To verify the effectiveness of the semantic segmentation module, we compare the model that directly utilizes the attention map of each object obtained from the image captioning model as the object mask.

\end{itemize}

For the last two baselines, we only compare them through qualitative visualization since they achieve far worse performance than the others. 

\subsubsection{Implementation Details}
Our SentiGAN holds a similar network structure as MUNIT \cite{huang2018multimodal}. It contains a content encoder, a style encoder, a decoder, and a discriminator. The content encoder contains two sub-encoders (image and object-oriented) that share the same weight, and the decoder includes two sub-decoders (image and object-oriented) that share the same weight. The content encoder consists of several strided convolutional layers to downsample the input and several residual blocks for further transformation. The style encoder consists of several strided convolutional layers, followed by a global average pooling layer and a fully connected layer. The decoder processes the content code by a set of residual blocks with Adaptive Instance Normalization to incorporate the style and content information. The output is further fed into several upsampling and convolutional layers to reconstruct the transferred image. For the training of SentiGAN, we set the hyper-parameters $\lambda_{1}$, $\lambda_{3}$ and $\lambda_{8}$ to 1, $\lambda_{2}$, $\lambda_{4}$ and $\lambda_{5}$ to 10, $\lambda_{6}$ and $\lambda_{7}$ to 0. We find that employing the object-level image reconstruction loss is sufficient for object-level supervision. As \cite{huang2018multimodal}, we use the Adam optimizer \cite{kingma2014adam} with $\beta_{1}$ = 0.5, $\beta_{2}$ = 0.999, and an initial learning rate of 0.0001 which decreased by half every 100,000 iterations. For the object mask extraction, we set the hyper-parameter $\alpha$ to 1.4, which performs the best to match the segmentation mask to the object.

\subsection{Experiment Results}
\subsubsection{Task 1}

\begin{table}[htbp]
\vspace{-2mm}
  \small
  \caption{\label{tab:result1} The coarse-level sentiment transfer performance of different models evaluated by the pre-trained sentiment classification model. The positive/negative rate represents the rate of predicting the positive/negative transferred images as positive/negative. The predicted positive/negative rate of the input images is listed in the first two rows. }
  \scalebox{0.8}{
  \begin{tabular}{|c|c|c|c|}
    \cline{1-4}
    &Positive Rate&Negative Rate&\\
    Input Images &0.540&0.460&\\ \cline{1-4}
      & True Positive Rate& True Negative Rate & Average\\ 
    MUNIT &0.582&0.478&0.530\\ \
    MUNIT + ObjSup &0.578&0.484&0.531\\ 
    MUNIT + ObjSup + CA &0.622&0.484&0.553\\ 
    SentiGAN - CA &0.594&0.502&0.548\\ 
    SentiGAN (IDL) &0.580&0.506&0.543\\ 
    SentiGAN &0.596&0.520&\textbf{0.558}\\ \cline{1-4}
    
    \end{tabular}}%
\vspace{-2mm}
\end{table}%
\vspace{-2mm}

\begin{table}[htbp]
  \small
  \caption{\label{tab:result2} The coarse-level sentiment transfer performance of SentiGAN evaluated by users. The hit rate represents the rate of selecting the positive transferred image as more positive in each positive-negative transferred image pair.}
  \scalebox{0.9}{
  \begin{tabular}{|c|c|c|}
    \cline{1-3}
      & Hit Rate& Miss Rate\\ \cline{1-3}
    User Study &0.724&0.276\\ \cline{1-3}
    \end{tabular}}%
\end{table}%

\begin{figure}[!t]
\vspace{-4mm}
\centering
\includegraphics[width=0.8\columnwidth]{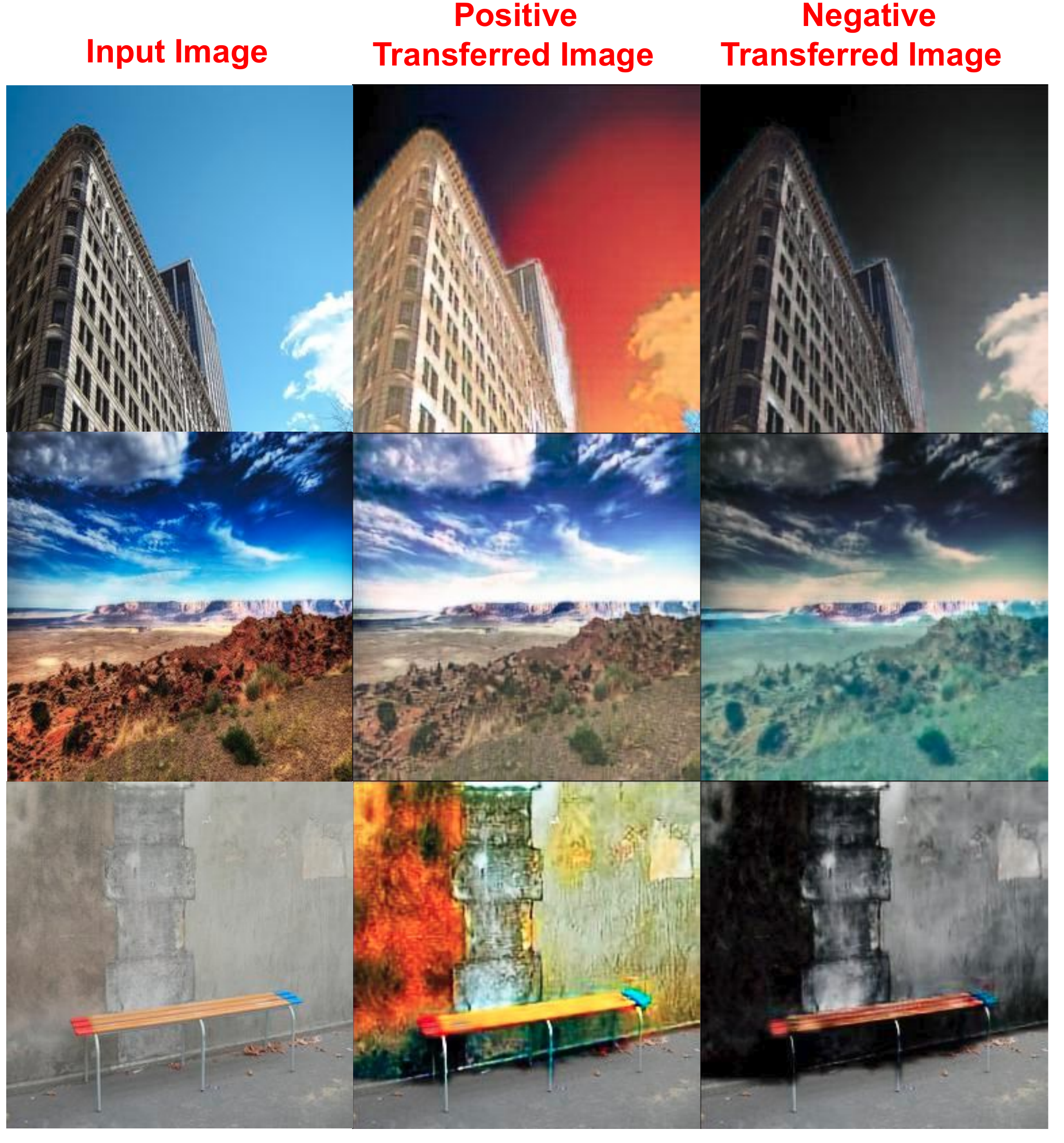}
\vspace{-3mm}
\caption{Example input images, the corresponding positive transferred images and negative transferred images.}
\label{fig:task1}
\vspace{-3mm}
\end{figure}

As described in Section~\ref{sec:eva}, for each input image, ten groups of positive reference images and ten groups of negative reference images are sampled to transfer the sentiment of the input one. The reference images are sampled by a pre-trained image sentiment classification model (based on ResNet-50) with a binary classification accuracy of 74.6\% on the original VSO test set. To evaluate different models' performance, we obtain the 500 transferred positive images, and 500 transferred negative images generated by different models. We further use the pre-trained sentiment classification model to predict the sentiment of each transferred image. As shown in Table~\ref{tab:result1}, SentiGAN achieves the highest average true positive and negative rates. In other words, compared with other models, there are more sentiment transfer cases agreed by the image sentiment classification model, which indicates the effectiveness of the SentiGAN to transfer the image's coarse-level sentiment. 

To further verify the sentiment transfer at the user level, for the 500 positive-negative transferred image pairs predicted by SentiGAN, we ask five volunteers to choose the more positive image of each pair with each volunteer responsible for 100 pairs. As shown in Table~\ref{tab:result2}, the rate of selecting the positive transferred image as more positive is 72.4\%, demonstrating that the transfer of sentiment can be commonly observed and appreciated by the users. Figure~\ref{fig:task1} shows several sentiment transfer cases produced by SentiGAN.

\subsubsection{Task 2}

\begin{table}[htbp]
  \small
  \caption{\label{tab:result3} Different transfer strategies evaluated by users. We show the rate of selecting the corresponding transferred images as the most real ones for each transfer strategy.}
  \scalebox{0.72}{
  \begin{tabular}{|c|c|c|c|}
    \cline{1-4}
      &Object-level Transfer& Global Transfer &Non-corresponding Object-level Transfer\\ \cline{1-4}
    User Study &0.672&0.288 &0.040\\ \cline{1-4}
    \end{tabular}}%
\end{table}%

\begin{figure*}[!t]
\vspace{-4mm}
\centering
\includegraphics[width=1.85\columnwidth]{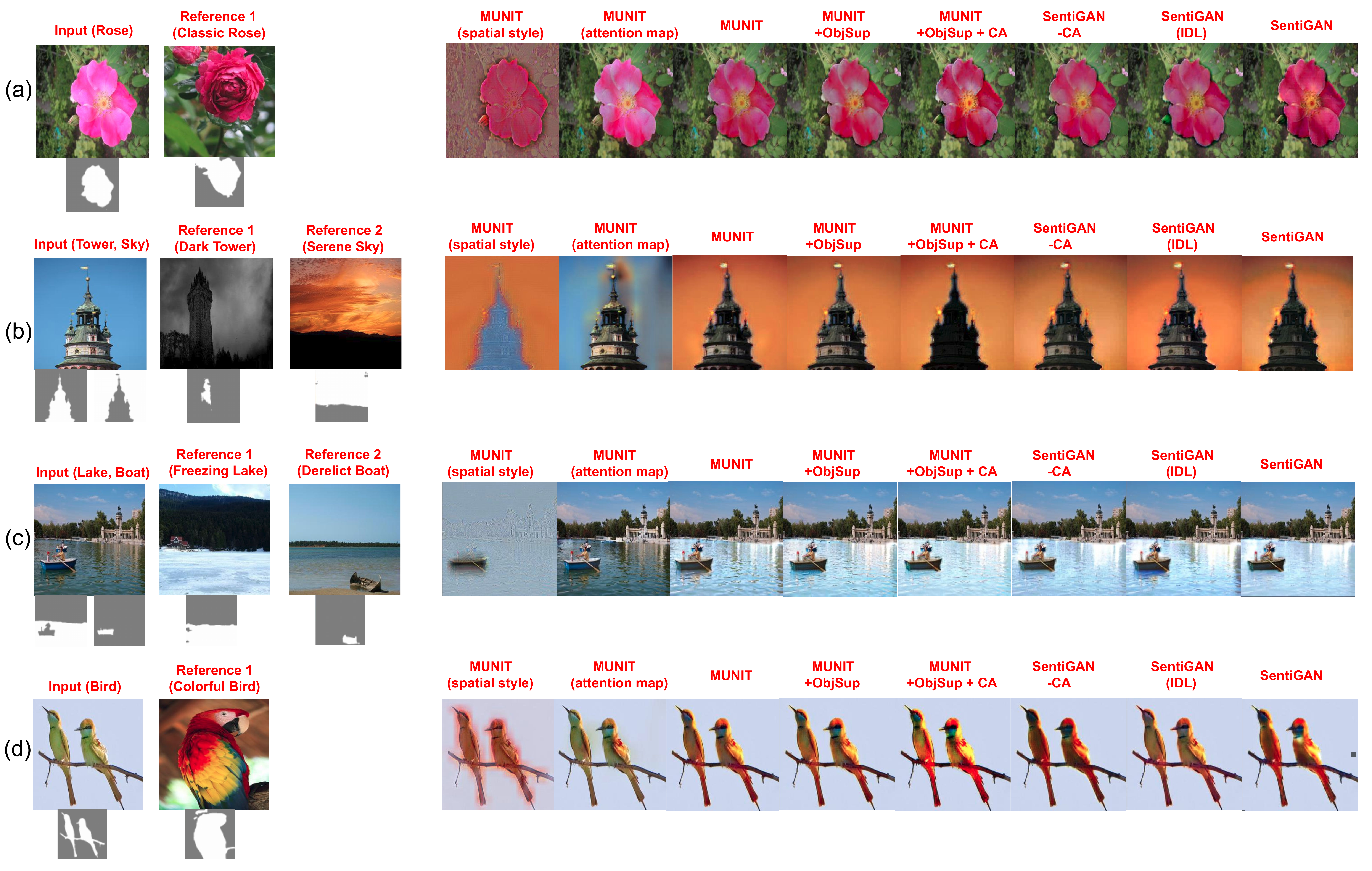}
\vspace{-3mm}
\caption{Example input images/object masks, reference images/ANPs/object masks, and the transferred images  by different models. }
\label{fig:task3}
\vspace{-5mm}
\end{figure*}
\begin{table}[htbp]
  \small
  \caption{\label{tab:result4} The sentiment consistency performance of different models evaluated by users. The hit rate represents the rate of the corresponding images selected as one of the most consistent with the reference images.}
  \scalebox{0.8}{
  \begin{tabular}{|c|c|}

    \cline{1-2}
    &Hit Rate \\\cline{1-2}
    MUNIT & 0.129\\ \
    MUNIT + ObjSup &0.150\\ 
    MUNIT + ObjSup + CA &0.189\\ 
    SentiGAN - CA &0.184\\ 
    SentiGAN (IDL) &0.123\\ 
    SentiGAN &\textbf{0.226}\\ \cline{1-2}
    
    \end{tabular}}

\end{table}%

\begin{figure}[!t]
\vspace{-1mm}
\centering
\includegraphics[width=0.8\columnwidth]{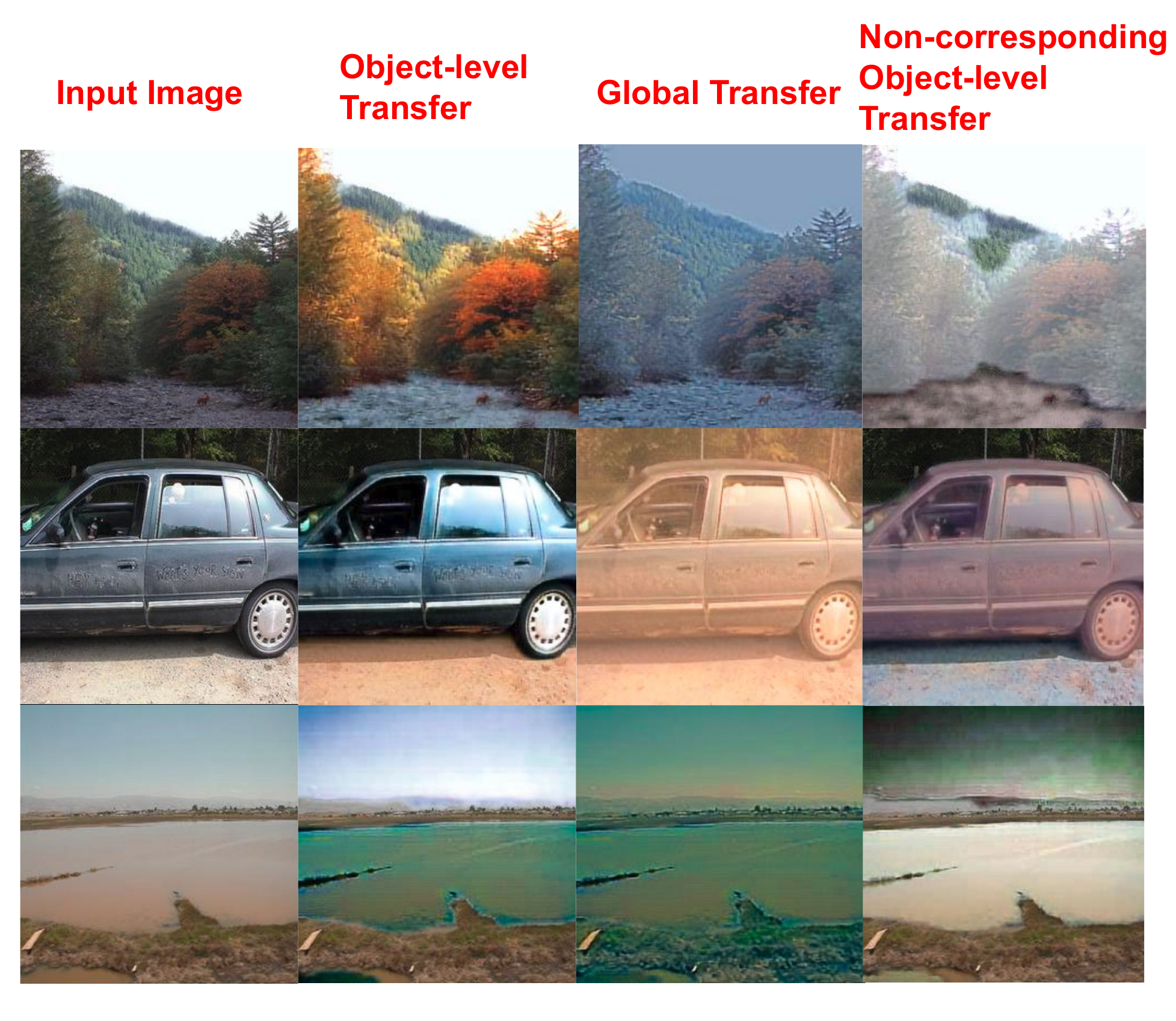}
\vspace{-5mm}
\caption{Example input images and the corresponding images transferred by different strategies.}
\label{fig:task2}
\vspace{-5mm}
\end{figure}

The second task verifies the effectiveness of transferring the image at the object level. As described in Section~\ref{sec:eva}, three types of transfer -- object-level transfer, global transfer, and object-level transfer with non-corresponding objects, are performed by SentiGAN to generate 50 groups of transferred images. For evaluation, we ask five volunteers to select the most real image for each group, with each volunteer responsible for 50 groups. As shown in Table~\ref{tab:result3}, for most groups, the volunteers agree that the image produced by the object-level sentiment transfer is the most real, which is consistent with the cases shown in Figure~\ref{fig:task2}.

\subsubsection{Task 3}

As described in Section~\ref{sec:eva}, the third task evaluates the sentiment consistency between the transferred image and the reference images. For each input image, we collect the transferred images predicted by different models and ask five volunteers to select one or multiple transferred images that are most consistent with the reference ones after letting them check both the reference images and the object masks. As shown in Table~\ref{tab:result4}, SentiGAN achieves the highest hit rate by a large margin, indicating the best performance in transferring the image sentiment from the reference images. Figure~\ref{fig:task3} illustrates several examples of the input images, reference images, and the corresponding transferred images predicted by different models. We can first observe that ``MUNIT (spatial style)'' and ``MUNIT (attention map)'' generate poor performance for the transfer. The former makes the images look petrified while the latter  encounters uneven transfer. Moreover, we find that our SentiGAN achieves better performance than the other models, especially in the aspect of color transfer. The rose of Figure~\ref{fig:task3}.(a), the tower of Figure~\ref{fig:task3}.(b), and the lake of Figure~\ref{fig:task3}.(c) transferred by SentiGAN hold more similar dominant color distribution to the reference objects than the others, enabling the transferred images to obtain similar sentiment to the reference images.

\section{Conclusions}

We study a brand new problem of image sentiment transfer and propose a two-step framework to transfer the image at the object level. The objects and the corresponding masks are first extracted by the combination of the image captioning model and semantic segmentation model. SentiGAN is further proposed to perform object-level sentiment transfer for the input objects. Evaluations based on the coarse-level sentiment, the realism, and the sentiment consistency of the transferred image have all demonstrated the effectiveness of the proposed framework. We plan to further improve the consistency of the transferred sentiment via language, such as imposing effective stylized image captioning supervision.

\bibliographystyle{ACM-Reference-Format}
\bibliography{acmart}

\end{document}